\documentclass[runningheads]{llncs}
\usepackage{graphicx}
\usepackage{enumitem}
\usepackage{comment}
\usepackage{cite}
\usepackage{amsmath,amssymb} 
\usepackage{color}
\usepackage{subcaption}
\usepackage{amsfonts}
\usepackage{multirow}
\usepackage{tabularx}
\usepackage{adjustbox}
\usepackage{mathtools}
\usepackage{amssymb}
\usepackage{soul}
\usepackage{algorithm}
\usepackage[noend]{algpseudocode}
\usepackage{booktabs}
\newcommand{\cmmnt}[1]{\ignorespaces}

\usepackage[belowskip=2pt,aboveskip=2pt]{caption}
 
\DeclareMathOperator*{\argmax}{arg max}

\def \bs {\boldsymbol}

\begin{document}
\pagestyle{headings}
\mainmatter
\def\ECCVSubNumber{100} 

\title{Towards Safer Self-Driving Through Great PAIN (Physically Adversarial Intelligent Networks)}

\titlerunning{Towards Safer Self-Driving Through Great PAIN}
%
\author{Piyush Gupta$^1$  \inst{*}\and
Demetris Coleman$^1$ \inst{*} \and
Joshua E. Siegel\inst{*}}
\authorrunning{P. Gupta et al.}
%
\institute{*Michigan State University, East Lansing MI 48824, USA\\
\email{\{guptapi1, colem404, jsiegel\}@msu.edu}}
\maketitle
\footnotetext[1]{Both authors contributed equally to this manuscript.}

\begin{abstract}
Automated vehicles' neural networks suffer from overfit, poor generalizability, and untrained edge cases due to limited data availability. Researchers synthesize randomized edge-case scenarios to assist in the training process, though simulation introduces potential for overfit to latent rules and features. Automating worst-case scenario generation could yield informative data for improving self driving. To this end, we introduce a ``Physically Adversarial Intelligent Network" (PAIN), wherein self-driving vehicles interact aggressively in the CARLA simulation environment. We train two agents, a protagonist and an adversary, using dueling double deep Q networks (DDDQNs) with prioritized experience replay. The coupled networks alternately seek-to-collide and to avoid collisions such that the ``defensive'' avoidance algorithm increases the mean-time-to-failure and distance traveled under non-hostile operating conditions. The trained protagonist becomes more resilient to environmental uncertainty and less prone to corner case failures resulting in collisions than the agent trained without an adversary.
\end{abstract}

\section{Introduction}
Automated Vehicles (AV's) are an imminent reality, and to reach main-stream adoption, AV's must ensure safe and efficient operation through intelligent decision making~\cite{kaur2018trust}. To this end, AV's must be exposed to and learn from an abundance of training data including real-world chaos~\cite{madrigal2017inside}. 

Due to economic cost constraints and the risk of physical damage, certain informative scenarios cannot be captured as real-world training data. Some self driving systems mitigate risk by avoiding high-speed operation in unfamiliar environments~\cite{kahn2017uncertainty}, or simulated environments may be used to allow AV's to observe atypical and infrequent experiences, providing faster-than-realtime exposure to simulated scenarios improving real-world performance. 

Researchers follow rigorous processes in collecting training data, generating scenarios and modeling vehicle dynamics. Segmentation and classification systems detect nearby objects, while perception systems estimate the trajectories for pedestrians and nearby vehicles and simulate edge cases that may lead to collisions. From these data, AV's neural networks learn features and typical responses but suffer limited (edge) data availability leading to model overfit and poor generalizability. As a result, operational vehicles exposed to unseen conditions – even slight variations on training data – may behave unexpectedly, with grave consequences. 

Simulators generate synthetic data, and augmentation tools increase data variability. However, simulations rely upon models with inherent biases, simplifications, or omissions. Specifically, traditional simulations implement scenarios designed by human users or with variational tools, leading to excluded edge cases. Researchers may utilize randomized scenarios to find edge cases from which to learn, but entropy is difficult to model. Human behavior, maintenance issues, and logical errors lead to a range of unpredictable behavior. These simulations lack the entropy of a chaotic environment which is critical to effectively train a \textit{defensive} self-driving car. To this end, we introduce a ``Physically Adversarial Intelligent Network (PAIN),'' pitting self-driving cars against one another to create a hostile, entropic environment. PAIN is based on ``Generative Adversarial Networks (GANs)~\cite{goodfellow2014generative},'' a means of pitting two neural networks, like those used to pilot self-driving cars~\cite{ghosh2016sad}, against one another to improve driving policies for both the attacker and the defender. 

In this paper, we pit a protagonist and an adversary agents against one another in the CARLA~\cite{Dosovitskiy17} simulation environment. The protagonist's objective is to drive safely from a start to a goal location, while the adversary seeks to maximize the damage to the protagonist. This helps generate real-world, noisy data where the protagonist learns to anticipate and avoid impending direct collisions, while the adversary learns to cause increasingly-unpredictable collisions. The resultant data are better-representative of real-world scenarios than that provided by pre-programmed or randomized simulation alone. 

Deep Reinforcement Learning (Deep RL) is a popular paradigm to train AV's (see~\cite{sallab2017deep} for an overview). Algorithms based on deep Q learning~\cite{li2017deep} and policy gradient methods~\cite{li2017deep} may be utilized to train agents with discrete and continuous action spaces, respectively. To ease implementation on constrained compute platforms, like those found in AVs, we selected the efficient Dueling Double Deep Q Network (DDDQN)~\cite{wang2015dueling} with prioritized experience replay~\cite{wang2015dueling} algorithm. We additionally utilize assisted exploration by incorporating a stochastic PID controller during exploration, and frame-skip~\cite{chen2019model} to accelerate training. We show that the protagonist trained with an adversary learns to drive defensively and exhibits higher success rate (safely reaching the goal location), with an increase in Mean-Time-to-Failure (MTTF) and average travel distance in unseen driving scenarios than the agent trained alone. We also show that with no surrounding vehicles, adversarial training does not impact performance negatively. 

 The major contributions of this work are fourfold:
 \begin{enumerate}[topsep=0pt, partopsep=0pt]
 \item We introduce PAIN, which pit self-driving vehicles with different objectives against one another with the environment-in-the-loop.
 \item We utilize the DDDQN algorithm with prioritized experience replay to train self-driving agents in a CARLA simulation environment.
 \item We utilize assisted exploration and frame-skip to speed up training of the coupled agents.
 \item We show that the trained ``defensive driving" agent becomes more resilient to edge cases than the agent trained without an adversary.
 \end{enumerate}
 Specifically, we show that the PAIN-trained avoidance algorithm outperforms the baseline (protagonist trained alone) in all measured performance metrics.
  
 This manuscript is structured as follows: in Section~\ref{Related Work}, we discuss prior art. In Section~\ref{Problem Formulation}, we formulate the problem, provide input representation and design reward functions for the two agents. We review Q-learning and traditional deep-Q learning algorithms in Section~\ref{Algo}. We discuss the DDDQN with prioritized experience replay, provide network architecture, and explore methods to accelerate training in Section~\ref{PAIN}. Simulation scenarios and results are provided in Section~\ref{Simulation}, and we conclude in Section~\ref{Conclusions} by identifying future opportunities and directions for continued work.
\section{Related Work}\label{Related Work}
\textit{Deep RL for automated driving:} Deep RL has demonstrated remarkable performance in sequential decision making tasks such as computer games~\cite{mnih2013playing} and robotic control~\cite{lillicrap2015continuous}. This growing popularity of Deep RL has made it popular for training AVs. \cite{sallab2017deep} reviews Deep RL approaches for automated driving and presents a framework leveraging Q-learning~\cite{watkins1992q}, Long-Short-Term-Memory (LSTM) networks~\cite{hochreiter1997long}, and Convolutional Neural Networks (CNN's)~\cite{krizhevsky2012imagenet}. End-to-end approaches such as imitation learning\cite{codevilla2018end,pan2018agile} have shown promising results. In~\cite{chen2019model}, the authors compare performance of model-free Deep RL algorithms such as DDQN~\cite{van2016deep}, twin delayed deep deterministic policy gradient (TD3)~\cite{fujimoto2018addressing}, and soft actor critic (SAC)~\cite{haarnoja2018soft} for automated urban driving. While the SAC performs best in urban autonomous driving, policy gradient methods (TD3 and SAC) are unsuitable for constrained compute platforms. DDDQN is a viable alternative for constrained platforms and improves upon challenges in traditional deep Q learning algorithms such as Q-value overestimation and unstable learning. 

\textit{Robust Adversarial Reinforcement Learning (RARL):} Many RL algorithms find a single policy that works on simulated data but may not adapt to the real world. The reality gap\cite{pinto2017robust} while transferring model domains is partly due to insufficient simulated data and limited scenario coverage leading to untrained edge cases.  \cite{pinto2017robust, pan2019risk} propose RARL as an extension to RL inspired by robust control methods. RARL uses two neural networks, a protagonist and an adversary, pitted against one another in a two player zero-sum game seeking to maximize and minimize a common reward function, respectively. The adversary acts as a destabilizing disturbance to the protagonist. In \cite{pan2019risk}, the two agents take turns performing actions to control the same vehicle so that the protagonist learns to robustly navigate a simulated race track. In RARL, the learning process involves alternately fixing one agent's policy while the other agent trains and then switching until both agents' policies converge. The trained protagonist learns to overcome disturbances while the adversary learns to produce more effective disturbances. Although the two neural networks control the same vehicle, the action space of the protagonist and the adversary are different. This allows the designer to model the actions of the adversary to match the expected disturbances that are encountered in the physical world. This forces the protagonist to learn robust policies capable of overcoming the realty gap.

\textit{Multi-agent reinforcement learning (MARL):} MARL~\cite{hu1998multiagent,bu2008comprehensive} extends RL to multi-agent systems by combining single-agent RL with game theory. MARL with competing agents can be viewed as an instance of RARL, where the adversary is an agent competing in the same environment. The adversary's goal is still to thwart the protagonist, but as an entity rather than a disturbance. Promising results have been shown for MARL in \cite{baker2019emergent}, where a game of virtual, team-based hide and seek led to complex behaviors. Like RARL, one team sought to maximize its reward, thereby minimizing the opposing team's reward and forcing both to learn new strategies to thwart one another. \cite{baker2019emergent} found that the learning algorithm may find and exploit faults in the simulation environment which amplifies the reality gap problem.

This approach can help generate data better-representative of the real-world through the PAIN framework. MARL will force the protagonist to learn robust policies under uncertainty, while the adversary will find novel strategies to crash, forcing the inclusion of edge cases in the training data.

\section{Problem Formulation}\label{Problem Formulation}
\begin{figure}
 \centering
 \includegraphics[width=0.6\linewidth, height=0.6\linewidth, keepaspectratio]{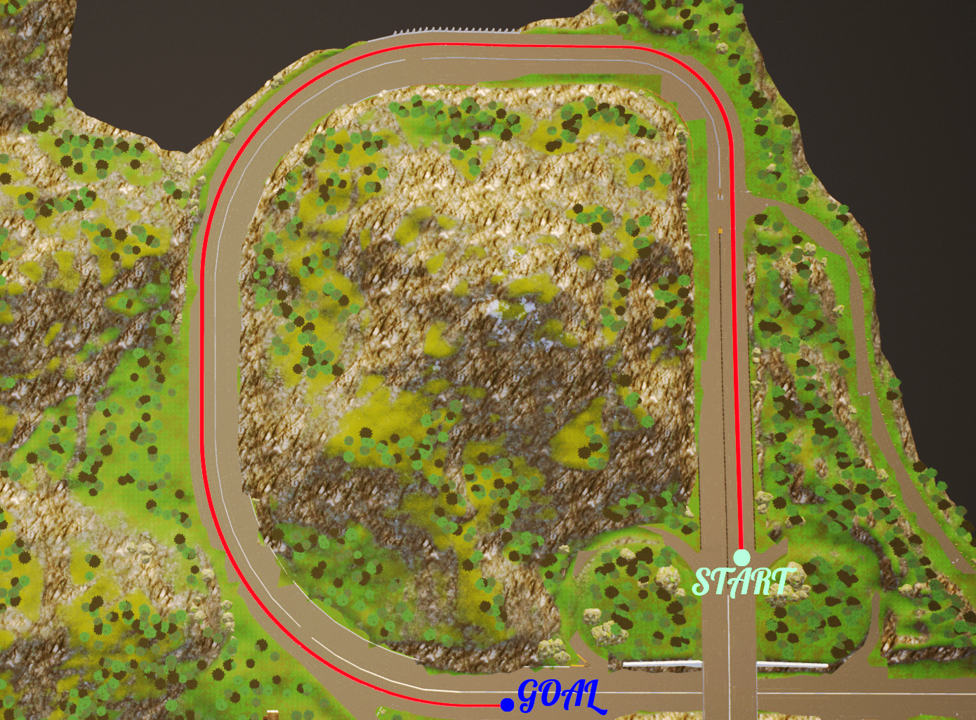}
 \caption{\footnotesize Bird-eye view of the track}
 \label{fig:route}
\end{figure}

 We consider two agents, an adversary and a protagonist, that seek-to-collide and to avoid collisions, respectively. The agents are trained using model-free Deep RL in the CARLA simulation environment (an open-source simulator for AV research). Figure~\ref{fig:route} shows the protagonist's $1.4 \ km$ desired route. 
The red curve shows the trajectory the trained protagonist should learn to drive safely from the start location (green circle) to the goal location (blue circle). We allow nine discrete actions $A \in \mathcal{A}$ for both agents, given by:
 \begin{equation}
 A=(T,S), \text{where}
 \end{equation}
$ T\in \{\textit{Constant, Accelerate, Decelerate}\} \text{ and } S \in \{ \textit{Constant}$, $\textit{Steer left, Steer right}\}$. For each action $A=(T,S)$, the control commands are calculated based on the previous control input:
\begin{subequations}\label{eq:control}
\begin{align}
St(St_{pr}, A) &= \begin{cases}
St_{pr}, & \text{if } S =
\textit{Constant},\\
\min\{St_{pr} + 0.2, 1\}, & \text{if } S = \textit{Steer right},\\
\max\{St_{pr} - 0.2, -1\}, & \text{if } S = \textit{Steer left},
\end{cases}\\
Th(Th_{pr}, A) &= \begin{cases}
Th_{pr}, & \text{ if } T =
\textit{Constant},\\
\min\{Th_{pr} + 0.2, 1\}, & \text{ if } T = \textit{Accelerate},\\
\max\{Th_{pr} - 0.2, 0\}, & \text{ if } T = \textit{Decelerate},
\end{cases}\\
Br(Br_{pr}, A) &= \begin{cases}
Br_{pr}, & \text{if } T =
\textit{Constant},\\
0, & \text{if } T = \textit{Accelerate},\\
\max\{Br_{pr} + 0.2, 1\}, & \text{if } T = \textit{Decelerate},
\end{cases}
\end{align}
\end{subequations}
 where $St \in [-1 \ 1], \ Th \in [0 \ 1], \ \text{and } Br \in [0 \ 1]$ are the steering, throttle and brake commands with subscript $pr$ denoting the control commands at previous time step. Incremental, control-targeted changes avoid abrupt changes in the outputs that might violate a real vehicle's kinematic constraints.
 We now provide our input representation and reward structures for the agents. 

\subsection{Input Representation}\label{sub:input}
\begin{figure}[!ht]
\centering
	\begin{subfigure}[b]{0.48\textwidth}
	 \centering
	 \includegraphics[width=0.95\linewidth, height=0.95\linewidth, keepaspectratio]{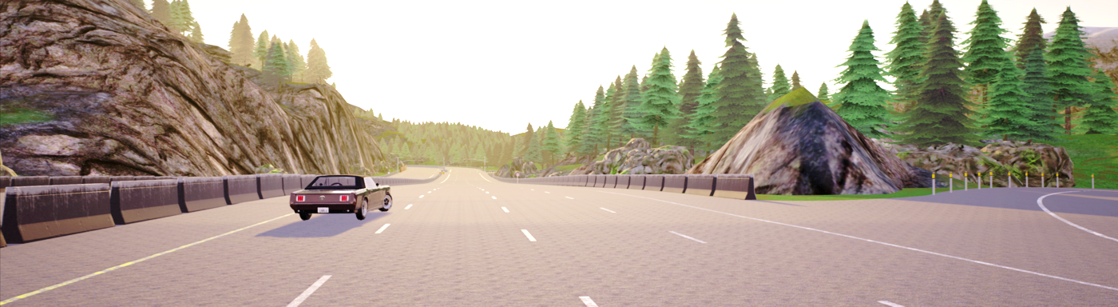}
 \caption{}
 \label{fig:rgb}
 \end{subfigure}
 ~
 \begin{subfigure}[b]{0.48\textwidth}
	 \centering
	 \includegraphics[width=0.95\linewidth, height=0.95\linewidth, keepaspectratio]{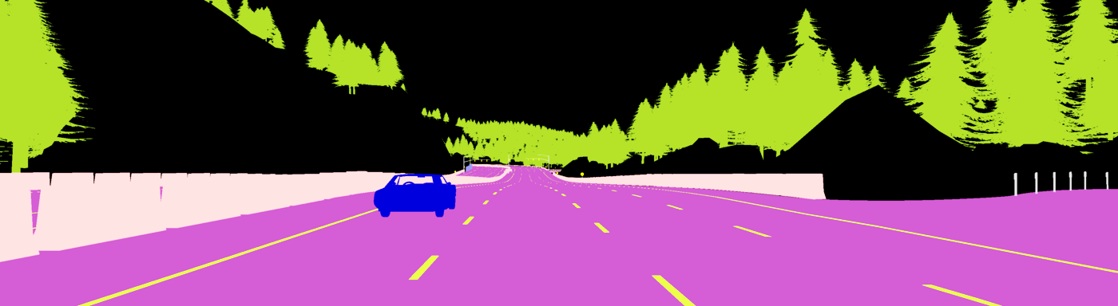}
 \caption{}
 \label{fig:semantic}
 \end{subfigure}
 ~
 \begin{subfigure}[b]{0.48\textwidth}
	 \centering
	 \includegraphics[width=0.95\linewidth, height=0.95\linewidth, keepaspectratio]{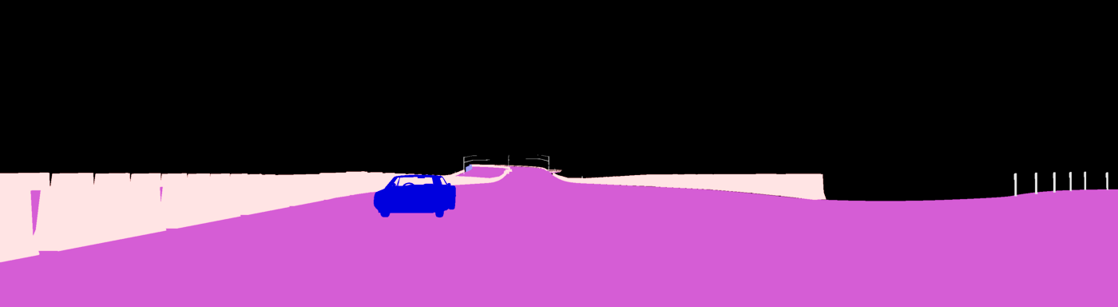}
 \caption{}
 \label{fig:processed}
 \end{subfigure}
 ~
 \begin{subfigure}[b]{0.48\textwidth}
	 \centering
	 \includegraphics[width=0.95\linewidth, height=0.95\linewidth, keepaspectratio]{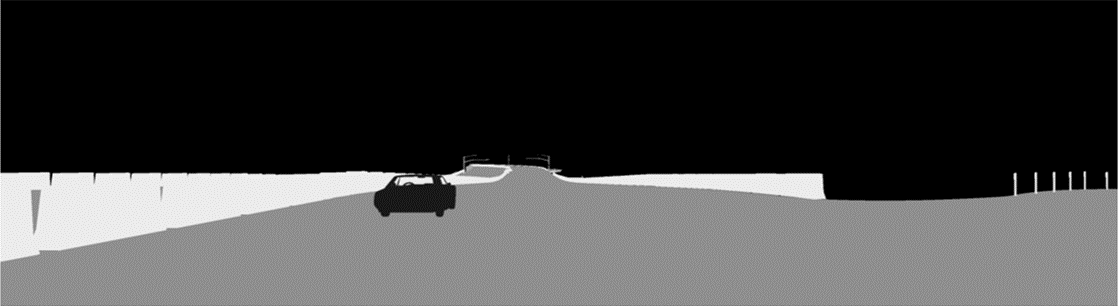}
 \caption{}
 \label{fig:gray}
 \end{subfigure}
 \caption{\footnotesize The visible car is an adversary seen by the progatonist's front view a) RGB front view, b) Semantic segmentation (CARLA segmented camera view), c) Post-processed, masked image, d) Normalized gray-scale image used as network input}
 \label{fig:input}
\end{figure}
The CARLA simulation environment includes high-dimensional information like road markings, traffic signs, objects, and weather effects. While AV's require these data for safe operation, the aim of this work is to study the impact of an adversary on training ``defensive'' agents and therefore does not require taking traffic laws into consideration. Unlike a real-world AV which would utilize RGB cameras, LiDAR, RADAR, and other sensors to cover the $360^{\circ}$ field-of-view of the vehicle, we instead utilize only the front view image, capturing $110^{\circ}$ field-of-view as perceptive input. 

To simplify the high-dimensional front view, we utilize CARLA's semantic camera to pre-segment an image.  Due to recent advances in semantic segmentation~\cite{he2017mask}, AVs can be equipped with such algorithms~\cite{ha2017mfnet}. We therefore assume the availability of such data. 

We further reduce the state complexity by post-processing the semantically segmented image to mask out vegetation and road markings. Figures~\ref{fig:rgb} and~\ref{fig:semantic} show a protagonist's front-view captured by (a) an RGB camera and (b) the semantic segmentation camera. The processed, masked image is shown in Figure~\ref{fig:processed}. This image is converted to a normalized gray-scale image (Figure~\ref{fig:gray}) of size 100x120, which is used in a sequence of four consecutive frames (to predict vehicle motion) as a perceptive input to the neural network. To simplify motion computation, we assume the availability of onboard sensors describing vehicle pose and relative motion. Therefore, our augmented input state $s \in \mathcal{S}$ to the neural network consists of:
\begin{enumerate}[topsep=0pt, partopsep=0pt]
 \item sequence of four segmented, masked, normalized grayscale forward images,
 \item vehicle motion state: $(v_{lon},v_{lat}, \omega, a_{lon},a_{lat})$ and,
 \item previous control commands: $(Th_{pr}, St_{pr}, Br_{pr})$, 
\end{enumerate}
where, $v_{lon} (a_{lon})$ and $v_{lat} (a_{lat})$ represents the longitudinal and lateral velocity (acceleration) of the ego vehicle, respectively, and $w$ represents its yaw rate. Vehicle motion state and control commands are easily accessible through common vehicle sensors such as IMU's, GNSS, encoders, etc.
\begin{figure}
 \centering
 \includegraphics[width=1\linewidth, height=1\linewidth, keepaspectratio]{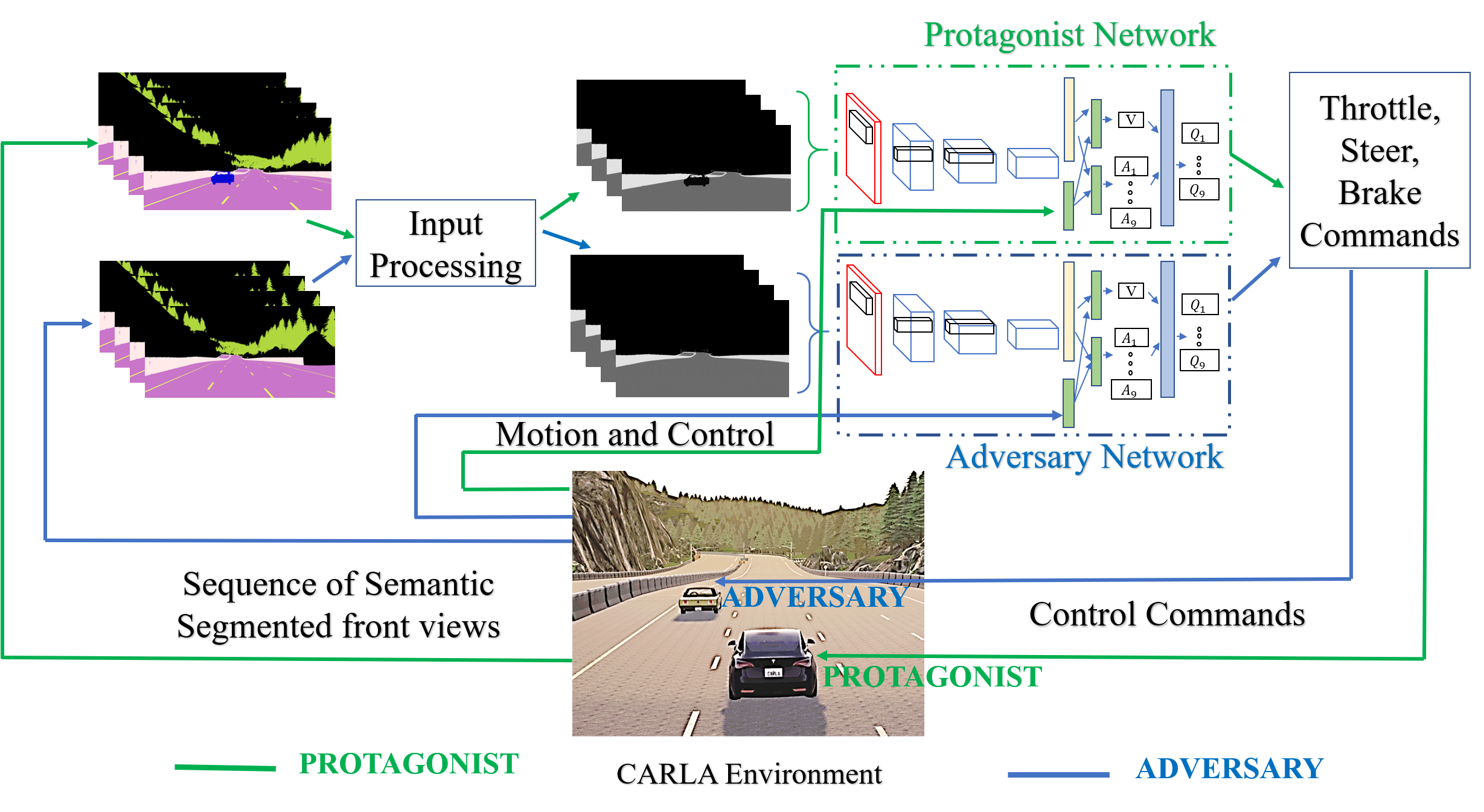}
 \caption{\footnotesize System framework with two coupled networks and environment-in-the-loop. Sequence of processed semantic segmented front view images along with vehicle motion and control states form a low-dimensional input for each agent, which then uses a deep neural network to estimate Q values, which help calculate control commands.}
 \label{fig:PAIN}
\end{figure}

Figure~\ref{fig:PAIN} shows the PAIN framework with the two coupled networks and the CARLA environment-in-the-loop.
By using low-dimensional input data, the trained algorithm will potentially be suitable for research and development on low-cost single board computers, eventually enabling PAIN to be trained on inexpensive physical test platforms\cite{siegel2019gamified}. 
\subsection{Reward Design}
\textbf{Protagonist:} The protagonist's objective is to safely drive from a start location to a goal location (Figure~\ref{fig:route}) in the minimum time without any collisions, subject to a maximum ``safe'' acceleration limit. This limit ensures the vehicle stays within the traction limits of the tire and suspension, as well as improves the occupancy experience for passengers by minimizing large and potentially-disruptive accelerations. For each state $s \in \mathcal{S}$ (Section~\ref{sub:input}) and action $A^P \in \mathcal{A}$ by the protagonist, we design the reward function $R^{P}(s,A^P) \in \mathbb{R}$:
\begin{equation}
 R^{P}(s,A^P)= R^{P}_v - R^{P}_a - R^{P}_{st} +R^{P}_{goal} -R^{P}_{col} - R^{P}_{cross},
\end{equation}
where $R^{P}_v$, $R^{P}_a$, $R^{P}_{st}$, $R^{P}_{col}$, $R^{P}_{goal}$, and $R^{P}_{cross}$ are the rewards based on absolute velocity $v^P$, absolute acceleration $a^P$, steering angle $st^P$, collision event, distance to goal $dis_{goal}$, and cross-track error $e_{cross}^P$ (minimum euclidean distance from protagonist location to its target trajectory), respectively. Reward terms are defined as following:
\begin{align}\label{eq:6}
 R^{P}_v &= \begin{cases}
-r_1, & \text{if } v^P \le v_{min},\\
r_2 \times \frac{v^P}{v_{max}}(cos(\theta^P)-sin(\theta^P)), & \text{if } v_{min} \le v^P \le v_{max},\\
0, & \text{if } v^P > v_{max},
\end{cases}
\end{align}
 \begin{equation}\label{eq:7}
 R^{P}_a= r_3\times \bs 1(a^P>= a_{max}),
 \end{equation}
 \begin{equation}\label{eq:8}
 R^{P}_{st}=r_4\times (st^P)^2,
 \end{equation}
  \begin{equation}\label{eq:9}
 R^{P}_{goal}=r_5 \times \left(1-\frac{dis_{goal}}{Route_{length}}\right) + r_6 \times \bs 1(dis_{goal}< \delta),
 \end{equation}
 \begin{equation}\label{eq:10}
 R^{P}_{col}=r_7 \times{\bs 1(collision^P)}, \\
 \end{equation}
 \begin{equation}\label{eq:11}
 R^{P}_{cross}=r_8 \times \bigg(\frac{2 \times e_{cross}^P}{road_{width}}\bigg)^2,
 \end{equation}
where $r_{i}$, $i \in \{1, \cdots, 8\}$ are positive constants, $\theta^P$ is the angle of the protagonist velocity with respect to the lane, and $\bs 1(\cdot)$ represents the indicator function which is true when the corresponding condition $(\cdot)$ is true. $v^Pcos(\theta^P)$ (-$v^Psin(\theta^P)$) encourages (penalizes) the velocity of the protagonist along (normal to) the direction of the lane. $v_{min}$, $v_{max}$ and $a_{max}$ are constants representing minimum and maximum limits of the absolute velocity and acceleration. The penalty $-r_1$ in $R^{P}_v$ minimizes unnecessary stopping. $R^{P}_a$ and $R^{P}_{st}$ penalizes the agent for large accelerations and over-steering, respectively, to improve stable and comfortable driving. $R^{P}_{col}$ provides a penalty for collisions determined by the CARLA collision detector, which provides information such as object ids and intensity for each occurrence. $R^{P}_{goal}$ provides a reward based on the distance to the goal normalized against the route length $Route_{length}$. It also provides a reward if the protagonist successfully reaches the goal point within a threshold $\delta$. $R^{P}_{cross}$ penalizes the vehicle based on the cross-track error from the desired path which we normalize with the half-width of the road $\frac{road_{width}}{2}$. 

Other reward functions were considered, including functions that replace $R_v^P$ and $R^P_{goal}$ with:
\begin{equation}
 R^{P}_v = \begin{cases}
-r_1, & \text{if } v^P \le v_{min},\\
r_2 \times \frac{v^P}{v_{max}}, & \text{if } v_{min} \le v^P \le v_{max},\\
0, & \text{if } v^P > v_{max},
\end{cases}
\end{equation}
\begin{equation}
R^{P}_{goal}=r_5 \times \left(1-\frac{dis_{subgoal}}{dis_{seg}}\right) + r_6 \times \bs 1(dis_{goal}< \delta),
\end{equation}
where we divided the protagonist target trajectory into segments of length $dis_{seg}$, with each segment end waypoint acting as an intermediate destination. Based on the segment closest to the protagonist, a reward based on the distance to the next sub-goal point $dis_{subgoal}$ encouraged it to follow the sub-goals to reach the final goal location. However, frequent transitions in sub-goal way-points with changing-segments lead to unstable driving at transition points due to sudden variation in $dis_{subgoal}$. Furthermore, this $R_v^P$ leads to perpetual collisions due to large velocity component normal to the lane. 
\eqref{eq:6} and \eqref{eq:9} performed better as a reward function.

\textbf{Adversary:} The adversary aims to maximize the peak absolute value of its acceleration by colliding with the protagonist. Due to limited front perceptive field of both the agents, we spawn the adversary facing towards the protagonist at some distance. We design the adversary's objective function to encourage driving towards the protagonist (without environmental collisions), making collision with the protagonist the top priority. For each state $s \in \mathcal{S}$ and action $A^A \in \mathcal{A}$ by the adversary, we utilize the reward function $R_{adv}(s,A^A) \in \mathbb{R}$:
\begin{equation}
 R^{A}(s,A^A)= R^{A}_v + R^{A}_{col} + R^{A}_{dis} -R^{A}_{cross} + R^{A}_{goal}- R^{A}_{st},
\end{equation}
where $R^{A}_v$, $R^{A}_{col}$, $R^{A}_{dis}$, and $R^{A}_{cross}$ are rewards based on the absolute velocity $v^A$, collision event, distance from the protagonist $dis_{pro}$, and cross-track error $e_{cross}^A$, respectively. To encourage driving towards the protagonist, we do online route planning from the adversary spawn location to the protagonist start point and utilize it to compute the cross-track error term $R^{A}_{cross}$. $R^{A}_{goal}$, and $R^{A}_{st}$  rewards are analogous to the protagonist's to encourage driving towards the goal point (starting point of protagonist) and limiting over-steering, respectively. Other terms are defined as:
\begin{align}
R^{P}_v &= \begin{cases}
r_2 \times \frac{v^A}{v_{max}}(cos(\theta^A)-sin(\theta^A)), & \text{if } 0 \le v^A \le v_{max},\\
0, & \text{if } v^A > v_{max},
\end{cases}
\end{align}
\begin{equation}
R^A_{col}=r_{10} \times {\bs 1(collision^A={pro})} -r_7 \times {\bs 1(collision^A\ne{pro})} ,
\end{equation}
\begin{equation}
 R^A_{dis}=r_{9} \times \left(1-\frac{dis_{pro}}{d}\right), 
 \end{equation}
 \begin{equation}
 \ \ R^{A}_{cross}=r_8 \times \bigg(\frac{2 \times e_{cross}^A}{road_{width}}\bigg)^2\times {\bs 1(e_{cross}^A > \Delta)}, 
\end{equation}
where $r_{9}$, $r_{10}$, $\Delta$ and $d$ are positive constants. $R^A_v$ provides reward to the adversary for driving at high velocity opposite to the protagonist, since doing so will produce peak maximum acceleration and maximum damage potential. $R_{col}^A$ provides a reward to the adversary in case of successful collision with the protagonist and penalizes it for colliding with the environment. $R_{dis}^A$ rewards the adversary based on its distance to the protagonist. The term $R^{A}_{cross}$ penalizes the adversary for large ($> \Delta$) cross-track error, making the adversary trajectory susceptible to environmental collisions. This avoids forcing any specific trajectory on the adversary. Note that without the reward terms $R^{A}_{cross}$, $R^{A}_{goal}$, and $R^{A}_{st}$, the adversary learns to circle in the roadway at $v_{max}$, blocking the protagonist's route. While such a policy will produce high hit rate for the adversary, the generated data will not be as diverse as that from other policies.

\section{Review of Deep Q Learning Algorithms}\label{Algo}

In RL, Q-learning~\cite{watkins1989learning} is often used to train agents with discrete action spaces. Deep Q-learning algorithms such as Deep Q Networks (DQN)~\cite{mnih2013playing} and Double Deep Q Networks (DDQNs)~\cite{van2016deep} utilize neural networks to better contend with large state spaces by estimating Q-values. In this section, we review Q-learning, followed by a discussion on common deep Q learning algorithms for discrete action spaces.

\subsubsection{Q-Learning:}
Q-learning is an off-policy model-free RL approach in which the agent \textit{estimates} the optimal Q-value for each action. For a given policy $\pi$, a \textit{true} Q-value (or action-value) for action $a$, $Q^{\pi}(s,a)$, is defined as the expected discounted reward for executing action $a$ at current state $s$ and following policy $\pi$ thereafter. Therefore,
\begin{equation}
Q^\pi(s, a) \equiv E [R_1 + \gamma R_2 + ... | S_0 = s, A_0 = a, \pi],
\end{equation}
where $\gamma \in [0,1]$ is a discount factor that represents the trade off between the importance of immediate and future rewards. The optimal Q-value is defined as 
$ Q^*(s,a) \equiv Q^{\pi^*}(s,a) = \max_\pi Q^\pi (s,a)$, such that  $V^*(s) = \max_a{Q^*(s, a)}$ is the optimal value in state $s$. Since an optimal policy is unknown during learning stage, the objective for any Q-learning based algorithm is to \textit{estimate} the optimal Q-values, $ Q^*(s,a)$, from which an optimal policy may be formed as $\pi^*(s) \equiv a^* = \argmax_a{Q^*(s, a)}$. Although there might exist multiple optimal policies $\pi^*$, the optimal Q-values are unique. 

To solve problems with large state-spaces, deep Q-learning methods approximate $Q^*(s,a)$ with a parameterized value function $Q(s, a; w)$. These parameters are updated after taking action $A_t$ in state $S_t$ and observing the immediate reward $R_{t+1}$ and resulting state $S_{t+1}$. This can be expressed as 
\begin{equation}
w_{t+1} = w_t+\alpha (Y^Q_t - Q(S_t, A_t; w_t))\nabla_{w_t} Q(S_t, A_t; w_t)
\label{eq:Q_update_1}
\end{equation}
where $\alpha$ is a step size and $Y^Q_t$ is the target Q-value defined as 
\begin{equation}
Y^Q_t \equiv R_{t+1} + \gamma \max_a Q(S_{t+1}, a; w_t) \label{eq:Q_update_2}
\end{equation}

We now review two commonly used deep Q learning algorithms: DQN's and DDQN's, which utilize neural networks to estimate the optimal Q-values.

\subsubsection{Deep Q Network (DQN):}
DQN utilizes a neural network that takes a state as an input and approximates the Q-values for each discrete action based on that state.
The network weights are updated using stochastic gradient descent resembling~\eqref{eq:Q_update_1} and~\eqref{eq:Q_update_2}. The original DQN algorithm suffers from drawbacks including slow convergence and unstable learning as the target Q-values change in conjunction with the weight updates. It also suffers from over-estimation of the Q-values due to lack of information about the Q-function in initial training stages. Choosing actions by maximizing potentially \textit{poorly-estimated} Q-values in these stages can lead to sub-optimal policies. DQN algorithm discards experiences after a single update, failing to learn from possibly rare and useful experiences. The streaming experiences make the updates strongly correlated which violates the i.i.d. assumption of many stochastic gradient-based algorithms. 

Two advances using target networks and experience replay have been proposed~\cite{mnih2015human} to improve the performance of the DQN algorithm. First, a target network $Q(s, a; w^-)$, with parameters $w^-$, is used to update the target Q-values $Y^{DQN}_t$ used by DQN by modifying~\eqref{eq:Q_update_2} to
\begin{equation}
Y^{DQN}_t \equiv R_{t+1} + \gamma \max_a  Q(S_{t+1}, a; w^-_t).
\end{equation}

Target network weights $w^-$ are copied from the online network weights $w$ every $\tau$ steps, and kept fixed during other steps. This speeds convergence during training as well as provides a more stable learning process. The second advance utilizes experience replay, in which past experiences are stored in a memory for a time and this memory is sampled uniformly to update the network. This makes weight updates less correlated and reuses past experiences to enhance training.

\subsubsection{Double Deep Q Network (DDQN):}
Double Deep Q-Networks \cite{van2016deep} improve upon the DQN algorithm by decoupling action selection from action evaluation, thereby reducing the likelihood of overestimation of Q-values. The update equation for DDQN weights is the same as that for DQN, with the exception that the target value $Y^{DQN}_t$ is replaced by $Y^{DDQN}_t$, defined as:
\begin{equation}
Y^{DDQN}_t \equiv R_{t+1} + \gamma Q(S_{t+1}, \argmax_a Q(S_{t+1}, a; w_t), w^-_t )
\end{equation}
where $w^-_t$ are the weights of the target network from the DQN. 

Although DDQN improves upon DQN, both methods fails to identify the difference between desirable and undesirable states, and suffer from overestimation of Q values and unstable learning. An automated vehicle trained with these algorithms might learn an unstable sub-optimal policy resulting in frequent failures including collisions. A Dueling, Double Deep Q Network (DDDQN)~\cite{wang2015dueling} improves upon these by estimating Q-values in a more reliable, faster, and stable manner. We illustrate this further in Section~\ref{PAIN}.

\section{Physically Adversarial Intelligent Network} \label{PAIN}

The PAIN framework trains coupled networks in a high-entropy environment to learn robust protagonist and adversary policies. We utilize DDDQN with prioritized experience replay for agent training. We now discuss the PAIN implementation's algorithms, architecture, and methods.

\subsection{Dueling Double Deep Q Network (DDDQN)}
In some states, the next action is critical. In others, it makes little difference. Whereas a action state may be critical for safety when driving on a narrow curvy road, it may not be in a wide open parking lot. For states with minimal action impact, it is unnecessary to estimate each action's value. To this end, DDDQN estimates Q-values by aggregating the state value $V(s)$ of being at a state and the advantage $A(s,a)$ of taking a specific action at that state. 
This is realized by splitting the output of the convolutional layers in a deep-Q network into two streams of fully connected layers, one for estimating value of the state $V(s)$ and the other for the action advantage $A(s,a)$.
The final Q-value estimate in DDDQN is obtained by aggregating estimates in an aggregation layer as following:
\begin{equation}\label{eq:DDDQN}
 Q(s,a;w,\alpha,\beta)=V(s;w,\beta)+(A(s,a;w,\alpha)-\frac{1}{|A|}A(s,a';w,\alpha)),
\end{equation} 
where $w,\alpha, \beta$ are the common network weights, advantage stream parameters, and the value stream parameters, respectively. 
Decoupling into two streams accelerates training by providing a more reliable estimate of Q values for each action~\cite{wang2015dueling}. Additionally, the network learns to identify whether a state is desirable while identifying the importance of each actions in that state. DDDQN therefore allows for faster, stable learning with reliable Q-value estimates.
\subsection{Prioritized Experience Replay (PER)}
PER~\cite{schaul2015prioritized} is an algorithm for sampling a batch of experiences from a memory buffer to train a network. It improves the policy learned by DQN algorithms by increasing the replay probability of experiences that have a high impact on the learning process. These experiences may be rare but informative. The prediction error of the Q-learning algorithm is used to assign a priority value $p_i$ for each experience stored in a memory buffer, which then generates a probability
\begin{equation}\label{eq:PER}
 \mathbb{P}(i) = \frac{p_i^\lambda}{\sum_k p_k^\lambda}
\end{equation}
by normalizing the priority values $p_i$ with the total priority values of all the experiences held in the memory. $\lambda \in [0, \ 1]$ is a hyper-parameter that adds randomness to experience selection. Sampling experiences from memory based on~\eqref{eq:PER} tends to add bias to the training data since high priority experiences will get selected more often. To remove this bias, PER weights experiences with importance sampling weights (IS) calculated as
\begin{equation}
 \omega_i = \Big(\frac{1}{N}\frac{1}{\mathbb{P}(i)}\Big)^\mu,
\end{equation}
where $N$ is the number of experiences in memory, and the hyper-parameter $\mu \in [0,1]$ controls the impact of IS weights on learning. The hyper-parameter $\mu$ is typically increased from a small value to $1$ throughout training because the weights are most important late in training (when Q-values start to converge). PER improves learning speed and policy quality when compared to uniform experience replay \cite{schaul2015prioritized}.

\subsection{Network Architecture}
\begin{figure}
 \centering
 \includegraphics[width=1\linewidth, height=1\linewidth, keepaspectratio]{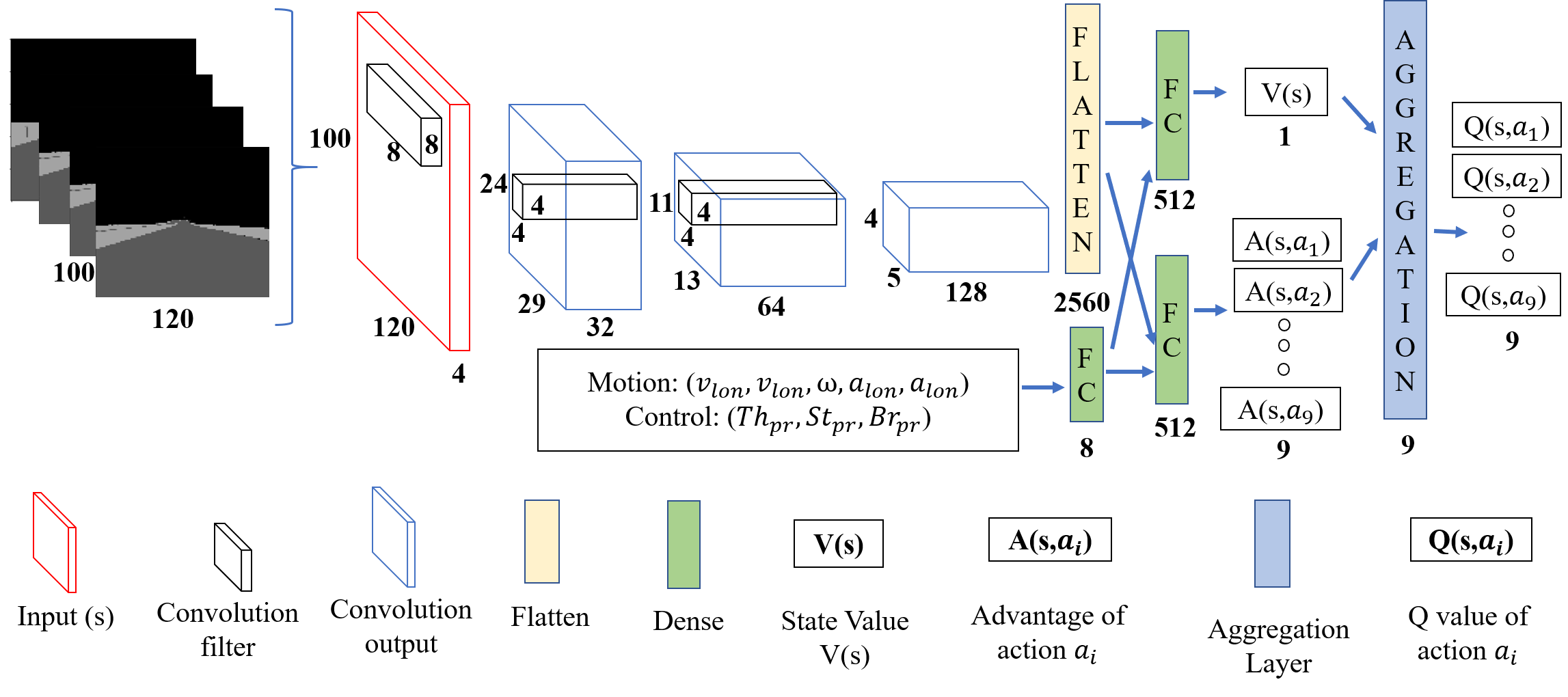}
 \caption{\footnotesize Neural Network architecture for both the agents.}
 \label{fig:architecture}
\end{figure}
Figure~\ref{fig:architecture} shows our DDDQN-based neural network architecture for both the agents. The sequence of four stacked normalized gray-scale images of size $100$x$120$ is passed through $3$ convolution layers with filter sizes 8x8x4-s-4, 4x4x32-s-2, and 4x4x64-s-2. The output after 3 convolutions (4x5x128) is flattened and passed to two 512-dimensional fully connected layers along with the vehicle motion state and previous control commands (see Section~\ref{sub:input}). The two fully connected layers estimate a one dimensional state value $V(s)$ and 9 dimensional advantage vector $A(s,a_i)$, $i \in \{1,2, \cdots, 9 \}$. Finally, the state value and advantage vector are aggregated (\eqref{eq:DDDQN}) through an aggregation layer to produce a $9$-dimensional vector corresponding to the Q-values for all possible actions. Similar architectures have been used to train agents to play computer games\cite{simonini2018introduction}.

\subsection{Assisted Exploration}
\begin{algorithm}[!ht]
 \caption{Assisted Exploration using PID controller}\label{assisted}
 \textbf{Input:} Set of all actions $\mathcal{A}$,\ PID action=$A_{PID} \in \mathcal{A}$, Exploration probability $\epsilon$, state $s$, DDDQN network $Net$, Previous Control command $(St_{pr}, \ Th_{pr}, \ Br_{pr})$\\\
 \textbf{Output:} Action $A$ and Control command $(St, \ Th, \ Br)$
 \begin{algorithmic}[1]
 \Procedure{GETACTION}{}
 \State $\textit{Sample a random number } \alpha \in [0 \ 1]$
 \If {$\alpha < \epsilon$}
 \State \textit{PID probability} $\mathbb{P}_{PID} \gets 0.75\epsilon$
 \State $\textit{Sample a random number } \alpha_{PID} \in [0 \ 1]$
 \If {$\alpha_{PID} < \mathbb{P}_{PID}$} 
 \State \textit{Action} $A \gets A_{PID}$
 \Else 
 \State \textit{Sample a random action} $A \in \mathcal{A}$
 \EndIf
 \Else
 \State \textit{Obtain network estimated Q values} $Q \gets Net(s)$
 \State \textit{Action} $A \gets \argmax_{a \in \mathcal{A}}\{Q\}$
 \EndIf
 \State $(St, Th, Br) \gets CONTROL((St_{pr}, \ Th_{pr}, \ Br_{pr}), A)$ \\
 \Return $A, (St, Th, Br)$
 \EndProcedure
 \end{algorithmic}
\end{algorithm}

Due to high-dimensional state space for autonomous driving, training is time-intensive. Researchers utilize techniques like imitation and transfer learning to speed up the training process\cite{siegel2019gamified}. We introduce assisted exploration utilizing a stochastic PID controller during exploration to speed up the training process. Specifically, we tune a PID controller to follow a target path and utilize it to generate preconditioning data along with the random exploration during training. Unlike the epsilon greedy approach, where random action are chosen during exploration, we choose exploration actions based on PID controller with probability $\mathbb{P}_{PID}$ and random actions with probability $1-\mathbb{P}_{PID}$. The probability of sampling action from PID controller $\mathbb{P}_{PID}$ is decreased with training steps to reduce the effect of PID controller over time. Algorithm~\ref{assisted} provides pseudo-code for the assisted exploration strategy. The input exploration probability $\epsilon$ exponentially decreases from its maximum $\epsilon_{max}$ to $\epsilon_{min}$ at a decay rate of $\Delta\epsilon$ with training steps $t_{steps}$:
\begin{equation}
 \epsilon= \epsilon_{min} + (\epsilon_{max} - \epsilon_{min}) * \exp(- \Delta\epsilon * t_{steps}),
\end{equation}
where $\epsilon_{min}=0.01$, $\epsilon_{max}=1$, and $ \Delta\epsilon = 0.000001$. Since the output of the PID controller does not necessarily satisfy~\eqref{eq:control}, the input PID action $A_{PID} \in \mathcal{A}$ is calculated by mapping the output of PID controller and previous control command to the corresponding action in $\mathcal{A}$. Without assisted exploration, the untrained agent may not explore enough of the state space to learn an effective policy from random actions. 

The function $CONTROL((St_{pr}, \ Th_{pr}, \ Br_{pr}), A)$ calculates the control input using~\eqref{eq:control}.

\subsection{Frame skip}
We utilize frame-skip technique to further accelerate training. In frame-skip, the action of the agent is kept unchanged for $k$ consecutive frames, after which a new action is applied for next $k$. This technique reduces training complexity and leads to stable policies. While training the protagonist, we change $k$ from one to three at pre-determined intervals. Initially actions are chosen from the PID controller with high probability. Utilizing frame skip and PID with large $k$ causes error accumulation and unstable agent motion, and frame-skip with random actions also leads to frequent collisions. Therefore, we increase $k$ from one to three after significant reduction in the exploration probability.

\section{Experimental Evaluation}\label{Simulation}
\subsection{Training and simulation scenario}
We first train the protagonist alone to learn a policy to reach the goal location without collisions. This model serves as a baseline for the protagonist. Once the baseline model learns to safety drive the route, we start our coupled training with the adversary. We clone the baseline policy to use as the initial policy for both the adversary and protagonist. We train two models for the protagonist; Model 1 ($243$K episodes) and Model 2 ($353$K episodes). 

\subsection{Performance Evaluation}
We utilize success rate, mean-time-to-failure (MTTF) and mean distance traveled (normalized with total route length) as performance metrics for the protagonist. The success rates are recorded at four checkpoints: 25\%, 50\%, 75\%, and 95\% of route length. If the protagonist successfully reaches the goal point, the MTTF corresponds to its track completion time. In case of failure through collision, we record the event's collision intensity (CI). 

We evaluate the protagonist's performance in four scenarios: (1) without any surrounding vehicles; (2) with 50 nearby vehicles driving in CARLA autopilot mode; (3) $5$ static vehicles obstructing the protagonist's route; and (4) against the adversary. Any vehicle initialized in autopilot mode randomly drives around the environment while respecting driving rules. We conduct 100 trials for each scenario and compare the performance with the baseline model of the protagonist trained without an adversary. 

\textit{Scenario 1: No surrounding vehicles}

We evaluate the performance of the protagonist in an environment with no other vehicles. This is the baseline agent's training environment, a useful scenario to evaluate whether the PAIN degraded baseline performance in safe environments (e.g. by learning over-conservative policies such as slow driving, or a bang–bang control strategy).

All tested models were able to complete the track with a $100\%$ success rate and therefore traveled the same distance without any collisions. The average reward of all the models were similar which indicates that the performance of the PAIN's protagonist does not deteriorate in safe environments. 

\textit{Scenario 2: Dynamic vehicles in autopilot mode}

This scenario emulates typical driving with surrounding vehicles. We randomly spawn 50 vehicles driving in autopilot mode in the environment. Due to the large environment, the agent might only interact with few vehicles (5-15) along its route. This scenario is a challenging as the protagonist has never seen more than one vehicle (the adversary) during training.

Some rear-end collisions occurred, which is unavoidable due to limited perceptive field ($110^{\circ}$ field-of-view) for all models' input. Table~\ref{tab:table-name1} compares the performance of the protagonist in Scenario $2$ with the baseline model. Protagonists trained through PAIN framework outperform the baseline in all performance metrics, and Model 2 outperforms Model 1 in all metrics, indicating that the PAIN framework will continue to improve over long timescales.
\begin{table}[!ht]
\begin{tabular}{@{}cccc@{}}
\toprule
\textbf{Model} & \textbf{Model 1} & \textbf{Model 2} \\ \midrule
\textbf{Average Reward} & 89.0\% & 107.6\% \\
\textbf{MTTF} & 38.0\% & 44.9\% \\
\textbf{Average Distance} & 29.6\% & 41.8\% \\
\textbf{Average CI} & -44.5\% & -65.5\% \\ \bottomrule
\end{tabular}
\quad
\begin{tabular}{@{}ccccc@{}}
\toprule
\textbf{Model} & \textbf{Baseline} & \textbf{Model 1} & \textbf{Model 2} \\ \midrule
\textbf{25\% Route} & 73\% & 86\% & 90\% \\
\textbf{50\% Route} & 58\% & 77\% & 87\% \\
\textbf{75\% Route} & 40\% & 65\% & 77\% \\
\textbf{95\% Route} & 23\% & 49\% & 61\% \\ \bottomrule
\end{tabular}
\caption{\label{tab:table-name1} \footnotesize Protagonist performance in Scenario 2. (Left table) Percentage change from the baseline model, (Right table) Success Rate}
\end{table}

\textit{Scenario 3: Static vehicles obstructing the protagonist route}

In this scenario, we obstruct the protagonist's trajectory by spawning $5$ static vehicles along its route. We choose random spawn locations for static vehicles from the waypoints encountered by the protagonist in absence of surrounding vehicles.
This provides edge cases where the static vehicles block the road and emulate real world accidents, roadkill, etc. It is a challenging scenario for the protagonist because: (i) multiple vehicles were not encountered in training, (ii) static vehicles were not in the training set, and (iii) the route blockage created by static vehicles forces the agent to deviate from its typical trajectory and attempt high-risk trajectories leading to unseen forwrad-facing camera data. Table~\ref{tab:table-name2} compares the performance of the protagonist models in Scenario $3$ with the baseline model. The protagonist models trained through PAIN framework outperforms the baseline in all performance metrics.
\begin{table}[!ht]
\begin{tabular}{@{}cccc@{}}
\toprule
\textbf{Model} & \textbf{Model 1} & \textbf{Model 2}\\ \midrule
\textbf{Average Reward} & 151.1\% & 259.3\% \\
\textbf{MTTF} & 37.0\% & 63.0\% \\
\textbf{Average Distance} & 31.9\% & 64.1\% \\
\textbf{Average CI} & -15.3\% & -27.9\% \\ \bottomrule
\end{tabular}
\quad
\begin{tabular}{@{}ccccc@{}}
\toprule
\textbf{Model} & \textbf{Baseline} & \textbf{Model 1} & \textbf{Model 2} \\ \midrule
\textbf{25\% Route} & 61\% & 79\% & 89\% \\
\textbf{50\% Route} & 20\% & 37\% & 59\% \\
\textbf{75\% Route} & 07\% & 22\% & 35\% \\
\textbf{95\% Route} & 04\% & 12\% & 22\% \\ \bottomrule
\end{tabular}
\caption{\label{tab:table-name2} \footnotesize Protagonist performance in Scenario 3. (Left table) Percentage change from the baseline model, (Right table) Success Rate}
\end{table}

\medskip
\medskip
 \textit{Scenario 4: Against Adversary }
 
 This scenario pits the protagonist against a trained adversary. The adversary was spawned facing towards the protagonist at some random distance in front of it. We evaluate the hit rate of the adversary, characterized as a successful adversary-protagonist collision of any force. \cmmnt{If the protagonist successfully gets past the adversary, it can easily complete the track due as there are no surrounding vehicles (Scenario 1).} In $100$ trials, the hit rate of the adversary against the baseline, Model $1$ and Model $2$ was found to be $22\%$, $34\%$, and $28\%$, respectively. Since the adversary is trained against the protagonist, it has a higher hit-rate against these models than the baseline suggesting that, like the protagonist, the adversary improves over episodes. In practice, we assume that vehicle collisions are incidental, and therefore the protagonist will be better-off in day-to-day driving, no matter how effective the adversary becomes. 

\section{Conclusions and Future Directions} \label{Conclusions}
In this work, we proposed a ``physically adversarial intelligent network (PAIN)" pitting multiple AVs against one another with the environment-in-the-loop. The coupled networks attempt to find faults in one another which improves the performance of both the protagonist and the adversary. We show that the protagonist  trained with the adversary outperforms the baseline model in all performance metrics. The presence of the adversary leads to more robust obstacle avoidance policies for the protagonist as well as provides edge case training scenarios that are difficult to pre-program.

There are several avenues for future research. We plan to extend our discrete action space to a continuous action space, utilizing state-of-the-art policy gradient methods such as soft actor critic (SAC), and twin delayed deep deterministic policy gradient (TD3). A continuous action space is a natural choice for the self-driving vehicles, and policy gradient algorithms such as SAC, and TD3 have been successfully used in automated driving~\cite{chen2019model}.

In our future work, we plan to add views from multiple cameras and other sensors to improve the perceptive field, utilizing transfer learning based on our current network trained with limited perceptive field to accelerate learning. 

Perhaps most important is the planned deployment of a PAIN on small-scale \textit{physical} vehicles\cite{siegel2019gamified}. It is of interest to transition these trained models from a simulation environment to scale-model vehicles in order to capture data stemming from the entropy inherent in physical systems that is difficult to model in simulations, even those based on the PAIN framework. Hundreds of small-scale vehicles can be built for the price of one full sized AV, leading to a happy balance of data collection cost, diversity, and speed. Using a physical platform will help cross the ``reality gap'' faster and more effectively than is feasible today.

\section{Acknowledgements}
We thank the NVIDIA Corporation for providing a Titan Xp and Vaibhav Srivastava for providing additional resources supporting this research.

%
%
\bibliographystyle{splncs04}
\bibliography{PAIN_arxiv}
\end{document}